# A Two-Stage Hybrid Model by Using Artificial Neural Networks as Feature Construction Algorithms

Yan Wang[1], Xuelei Sherry Ni[2], and Brian Stone[3]

[1]Graduate College, Kennesaw State University, Kennesaw, USA
[2]Department of Statistics and Analytical Sciences, Kennesaw State University, Kennesaw, USA
[3]Atlanticus Services Corporation, Atlanta, USA

## ABSTRACT

*We propose a two-stage hybrid approach with neural networks as the new feature construction algorithms for bankcard response classifications. The hybrid model uses a very simpleneural network structure as the new feature construction tool in the firststage, thenthe newly created features are used asthe additional input variables in logistic regression in the second stage. The modelis compared with the traditional one-stage model in credit customer response classification. It is observed that the proposed two-stage model outperforms the one-stage model in terms of accuracy, the area under ROC curve, andKS statistic. By creating new features with theneural network technique, the underlying nonlinear relationships between variables are identified. Furthermore, by using a verysimple neural network structure, the model could overcome the drawbacks of neural networks interms of its long training time, complex topology, and limited interpretability.*

## 1. INTRODUCTION

Recently, more and more financial institutions have extensively explored better strategies for decision making through the help of bank card response models. It is because theinappropriatecreditdecisioncouldresultinthedecliningprofitability of the marketing campaigns as well as huge amount of losses. After careful review of the literatures, it can be concluded that linear discriminant analysis (LDA) and logistic regression are the two widely used statistical techniques in bankcard response models [1] [2]. LDA requires the assumption that the linear relationship between dependent and independent variables, which seldom holds in most real datasets[3]. Further more, LDA is very sensitive to deviations from the multivariate normality assumption. On the other hand, logistic regression, which is designed to predict dichotomous out comes, does not require the multi-variate normality assumption. Moreover, logistic regression is shown to be more efficient and accurate than LDA under non-normality situations [4].Therefore, logistic regression has been acted as a good alternative to LDA for along time in bank ruptcy prediction, market segmentation, customer behaviour classification, and credit scoring modeling.

However, similar to LDA, logistic regression only explores the linear relationship among the independent variables and hence are reported to produce poor bankcard response capabilities in some cases [5]. As a result, neural network is increasingly found to be useful in modeling the bankcard response problems and are shown to outperform the logistic regression, since the neural network approach can identify subtle functional relationships among variables [6]. Furthermore, the neural network is particularly preferred in the situations where the variables exhibit complex non-linear relationships [7]. Even though neural network has the above-mentioned advantages, it is being criticized for its long training process, difficult to identify the relative importance of variables, and limited interpretability [8]. These drawbacks have limited the applicability in handling general bankcard response problems [9].

It is worth mentioning that, most research and application about neural networks focuses on using it as a modeling tool for classification problems. There is seldom research that uses this technique as a feature construction tool. Focusing on overcoming the cons of neural networks in bankcard response modeling including the long training time and the non-interpretability while focusing on taking advantage of the pros of neural networks including the exploration of non-linear relationships among variables, the authors believe that neural network should be a good supporting tool for logistic regression in terms of new feature constructions [8]. Thus, we will propose a two-stage hybrid approach in this study. By using simple neural network structures for feature constructions, we can explain the relationships among variables and avoid the long training time. In the meanwhile, the newly created features should be useful in improving the overall model performance.

The rest of the paper is organized as follows. Since the bankcard response model is used as an illustration in this paper, we will firstly review its related work in Section 2. Section 3 provides a detailed description about our model and its application on bankcard response classification, including the dataset description, the data pre-processing, the development of the one-stage model, the two-stage model, and the performance evaluation. The experimental results and the discussions are elaborated in Section 4. It is worth to mention that the descriptions and results in Sections 3 and 4 are based on Atlantic us data (credit card customer response dataset provided by Atlantic us Services Corporation). Then in Section 5, a public HMEQ data in [10] and also in the SAMPSIO library of SAS is used to further evaluate the consistency and reliability of the two-stage model. Finally, Section 6 addresses the conclusion and the future research directions.

## 2. RELATED WORK

The literature of commonly used techniques in bank card response modeling and credit scoring modelling are reviewed in this section. Based on these reviews, we will introduce the motivations of our study.

### 2.1. LOGISTIC REGRESSION

Logistic regression is one of the most widely used techniques in building credit scoring models and bankcard response models. The objective is to determine the conditional probability of a specific customer belonging to a class given the values of the independent variables of that observation by an equation of the form in (1), where $p$ is the probability of the conditional probability of a specific customer belonging to a class, $\beta_0$ is the intercept term, and $\beta_i$ is the $\beta$ coefficient associated with the independent variable $x_i$.

$$\log\left(\frac{p}{1-p}\right) = \beta_0 + \beta_1 * x_1 + \beta_2 * x_2 + \cdots + \beta_k * x_k \tag{1}$$

Since the $\beta$ coefficients could easily be converted into the corresponding odds ratios, one can easily interpret the magnitude of the importance of a certain predictor [11]. In addition, the criteria for assessing "goodness of fit" of logisticregressions such as the Hosmer-Lemes how statistic are widely accepted [12]. Furthermore, logistic regression is shown to be as accurate as many other techniques such as support vector machine when building the dichotomous outcomes [13]. Thus, in many financial institutions, logistic regression is the only acceptable tool for credit risk modeling and bankcard response modeling due to the regulations in financial industry.

## 2.2. NEURAL NETWORKS

Researchers aim at exploring advanced methodologies for bankcard response modeling to improve the performance. Neural networks have similar goal as in logistic regression and they aim at predicting an outcome based on the values of predictors. Compared with logistic regression, they could model any arbitrarily complex nonlinear relationships between independent and dependent variables as well as detect all possible interactions between predictors. Neural networks have successfully been used in a few studies for bankcard or credit modeling tasks. A neural network ensemble approach was applied for the bankcard response problem in [7]. In [14], a two-stage hybrid credit modeling was proposed by using neural networks and multivariate adaptive regression splines. Furthermore, a functional link neural network was implemented for bank credit risk assessment [15]. Thus, neural networks may represent an attractive alternative tologisticregressionif noregulationrestrictions.

On the other hand, however, neural networks are being criticized for their disadvantages. A neural network model is a relative "black box" in comparison to a logistic regression model. It has limited ability to magnitude the relative importance of a certain predictor and cannot easily determine which variables are the most important contributors to a particular output [16]. And there are no well-established criteria for interpreting the weights or coefficients in the network structure. Furthermore, the training time is long before a network model converging to an optimum learning state when the dataset is relatively large [17]. In addition, it is not easy to identify the optimal network's topology since model developers need to go through an empirical process to determine many training parameters such as learning rate, number of hidden nodes, and number of hidden layers [18]. As a result, in many financial institutions, neural networks have very limited applicability as the modelingtools.

Considering the pros and cons of neural networks, we propose, in this paper, to use simple neural network structures create new features, which can help improve the model performance but not cost too much time. In addition, the simple structure will make the interpretation a doable job.

## 2.3. FEATURE CONSTRUCTION ALGORITHMS

The main goal of feature constructions is to get a new feature which represents the patterns of the given dataset in a simpler way and hence makes the classification or prediction tasks easier and more accurate [19]. The widely used and well-known approaches include some generic feature construction algorithms such as $k$-means clustering, Singular Value Decomposition(SVD), and Principal Component Analysis (PCA). These algorithms create new features mainly focus on transforming the data and reducing the dimensionality [20]. For $k$-means clustering, the intuition for new feature constructions is to replace a group of similar features by a single representative feature [21]. SVD generates a new feature space in which individual features are linear combinations of features from the original space [22]. Similarly, PCA creates new features using a set of new orthogonal variables called principal components to display the important information from the datasets[23].

However, all the above-mentioned feature construction algorithms are un-supervised. That is, they do not consider the relationship between the input variables and the outputs at all. They can help reduce the dimensionality but the newly created feature may not be very useful in predicting the outputs. Furthermore, without kernel extension, those methods can only make linear summaries of the predictors. On the other hand, the neural network algorithm, as a supervised learning technology, could help generate new features that have a high predictability on the response and in addition, explore the non-linear relationship.

## 3. THE HYBRID MODEL AND ITS APPLICATION ON BANKCARD RESPONSE CLASSIFICATION BASED ON ATLANTIC US DATA

### 3.1. DATA DESCRIPTION

In order to access the feasibility and the effectiveness of the proposed two-stage hybrid model by using neural networks as feature construction tools, a dataset provided by Atlantic us Service Corporation was used here. We appreciate their sponsorship on this study so that we have the opportunity to evaluate our model based on the recent (2016) credit records. The dataset includes the records of 12,498 customers, and 538 features that are related with the customers' credit information. The target variable RESP_DV denotes a binary problem and can be defined as follows: 1 and 0 denotes customers with and without response after receiving the promotions of credit card, respectively. The ratio of customers with response is 80.01%. All theindependentvariablesarecontinuous.

### 3.2. DATA PRE-PROCESSING

To use the dataset in this study, some data pre-processing methods are applied for data cleaning and preparation. These methods are listed in the sequential order as follows:

(1) Replace invalid values in the data set using missing values.

(2) Randomly split the entire dataset into 60% training set and 40% validation set by using stratified random sampling method. The target variable is used as the stratification variable.

(3) Impute missing values with median and generate missing indicators as additional predictors.

(4) Conduct hierarchical variable clustering [24]. This method is applied before modeling to eliminate redundant features in the original data. Variables with the lowest $1 - R^2 ratio$ defined in (2) in each cluster is selected as the representative of the current cluster. That is, the variable that has the strongest linear relationship with the variables within the group, and the least relationship with the variables outside the group, would be chosen as the representative of the current cluster. Number of clusters are determined to preserve at least 90% of the data variability.

$$1 - R^2 ratio = \frac{1 - R^2_{own\_cluster}}{1 - R^2_{next\_closest\_cluster}} \tag{2}$$

(5) Transform all the variables with the WeightofEvidence (WOE) method [25].This is the standard approach in credit scoring. The transformation will encode variables in a few buckets, making the final log_reg coefficients $\beta_i$ from logistic regression interpretable.

After data pre-processing, 178 independent variables were selected for the final experiment. In addition, the training set has 7,499 records while the validation set has 4,999 records.

### 3.3. STAGE ONE OF THE HYBRID MODEL – NEURAL NETWORKS FOR NEW FEATURE CONSTRUCTION

In this section, the first-stage of the proposed hybrid model, which aim satusing neural networks as feature construction tools, is described. Figure1shows the block diagram of using neural networks for new feature constructions. It containssix steps labelled from A to F. Step D is implemented in Python (version 3.5) due to its computational ability and there maining five steps are applied in SAS Enterprise Guide (version12).

As stated in Section 3.2, 178 independent variables remain after variable clustering. These variables form15,753 possible pairs of variables by using the *n-choose-k* combination described in (3), where *n*and *k*are valued 178 and 2, respectively.

$$C(n,k) = \frac{n!}{(n-k)!\,k!} \tag{3}$$

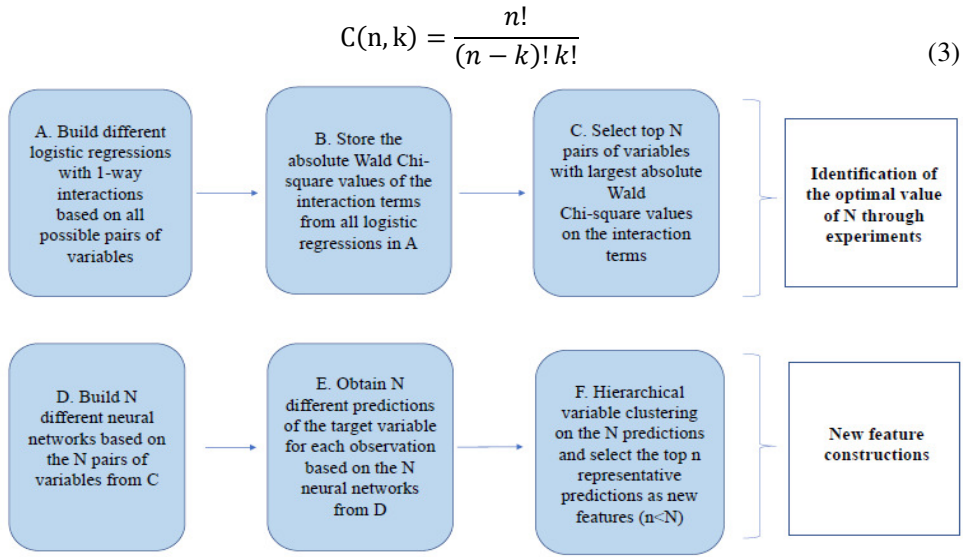

Figure 1. The block diagram of using neural networks for new feature constructions

Therefore, in step A of Figure1, 15,753 different logisticregressionswith1-way interaction would be build based on these 15,753 pairs of variables. Based on these15,753 logistic regressions, Wald Chi-square tests are individually implemented to test the significance of the interaction terms. Take a certain pair of variables containing variables *AMS3726* and *AMS3161* as an example. The format of the logistic regression built in step A would be defined in equation (4), where *p* denotes the probability of respondents (i.e., *RESP_DV* = 1), *AMS3726* denotes number of open bankcard accounts with update within 3 months, *AMS3161* denotes total balance of open bank card accounts with update with in 3 months, and *AMS3726*AMS3161* denotes the interaction term of the two variables. Then, the absolute Wald Chi-square value of *AMS3726*AMS3161*(or, the corresponding *p* value) is recorded and stored in Step B. Although there are 15,753 iterations for steps A and B, it takes only about2 hours in SAS by using the computer with 3.3 GHz Intel Core i7 processor for our study since the format of the logistic regression is relatively simple.

$$\log\left(\frac{p}{1-p}\right) = \beta_0 + \beta_1 * AMS3726 + \beta_2 * AMS3161$$
$$+ \beta_3 * AMS3726 * AMS3161 \tag{4}$$

In step C, the top *N* pairs of variables with highest absolute Wald Chi-square values (corresponds to lowest *p* values)from their interaction terms in the logistic regressions are selected. In this paper, the value of *N* is set to50 via experiments. We have tried to set *N* to be 25, 50, 100, and 150 in our study. Results show that when *N* exceeds 50, there is no large improvement in the final model performance. Considering that the training time in step D increases as N becomes larger, we set the value of N to be50.The run-time experiment shows that it takes only about 20 seconds in Python to build the 50 different neural networks. Because of using different datasets in the credit response problems, it would be better for future researchers to try several different values of N for obtaining a satisfying classification performance with a relatively short training time.

In step D, the selected 50 pairs of variables are used to construct 50 different neural networks on the training set. Consider the pair of variables containing *AMS3726* and *AMS3161*again for the illustrative purpose. The built neural network structure is shown in Figure 2. There are two input nodes in the input layer, denoting two input variables. The number of hidden layer is set to one since we do not want to create new featuresthat are constructed based on too complex relationships between the two input variables. The output node calculates the predicted probabilities of the responsive status of the customers (i.e., with response or without response) in this study. The activation functions used in the hidden and output layers are both sigmoid defined in equation (5).

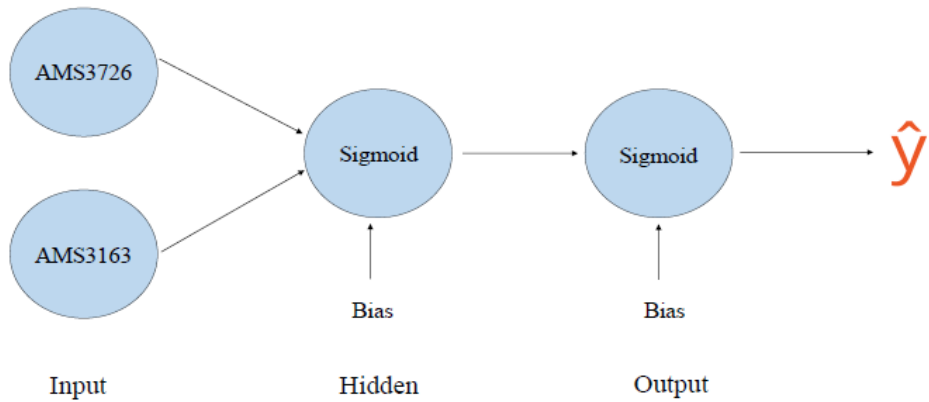

Figure 2. The simple neural network structure used for feature construction.

$$\text{sigmoid}(x) = \frac{1}{1 + \exp(-x)} = \frac{\exp(x)}{1 + \exp(x)} \qquad (5)$$

For setting appropriate number of hidden nodes, the trial and error approach with the range from one to five neurons is used. As a result, there are no significant difference of the model performance when changing the number of hidden nodes in the hidden layer. Therefore, we set the number of hidden nodes as one to keep the simplicity of the neural network structure. The training of a network is implemented with various learning rates ranging from 0.00001 to 0.1 and traininglengthsrangingfrom100to10,000iterationsuntilthenetworkconverges. The settings of above hyper-parameters ensure the converge of the neural network within a relative short time (<3 minutes on the computer with 3.3 GHz Intel Core i7 processor).

Instep E, for each observation in the training set, there would be 50 different predicted probabilities of respondingcalculated from the 50 different neural networks in step D. These predicted probabilities are denoted as $\hat{y}_0$, $\hat{y}_1$, ...., $\hat{y}_{49}$.They will enter the hierarchical variableclusteringanalysisinstepFtoreducethe potentialmulticollinearityissue. The parameter settings in step F are the same as those in the clustering analysis described in Section 3.2. The clusterrepresentativesareconsideredasthenewly constructed features by using neural networkalgorithms. In our application, there are 22 newly created features being identified as cluster representatives. They will be added into the model as additional predictors in stage two.

### 3.4. STAGE TWO OF THE HYBRID MODEL – LOGISTIC REGRESSION

In the second stage, logistic regression with newly created features by using the neural network algorithm was built following the steps illustrated in Figure 3.In Figure 3, the modeling procedure starts from the block coloured with red and contains four main steps as follows:

(1) Initial modeling. Features (without creating new features for one-stage modelandwithnewlycreatedfeaturesfortwo-stagemodel)areusedtodevelop the logistic regression model by applying the stepwise selection method. The significant levels to enter and leave the model are used as the default values in the stepwise selection procedure in SAS (i.e., 0.15) to reduce the possibility of excluding the potentially significant variables as well as of including too many insignificant variables. The model is generated on the training set and scored on the validation set.

(2) Checking variance of inflation factor (VIF). Variables selected by the logistic regression would be used to calculate VIF values in multiple linear regression models. Variables with VIF larger than 10 are considered to have potential multicollinearity problems and would be removed[26].

(3) Checkingthevariablecoefficients.AsdescribedinSection3.2,the variables are transformed to their WOE values. Theoretically, the relationship between the WOE- formed variables and the target variable should be positive [27]. Therefore, variables with negative coefficients are removed from the model.

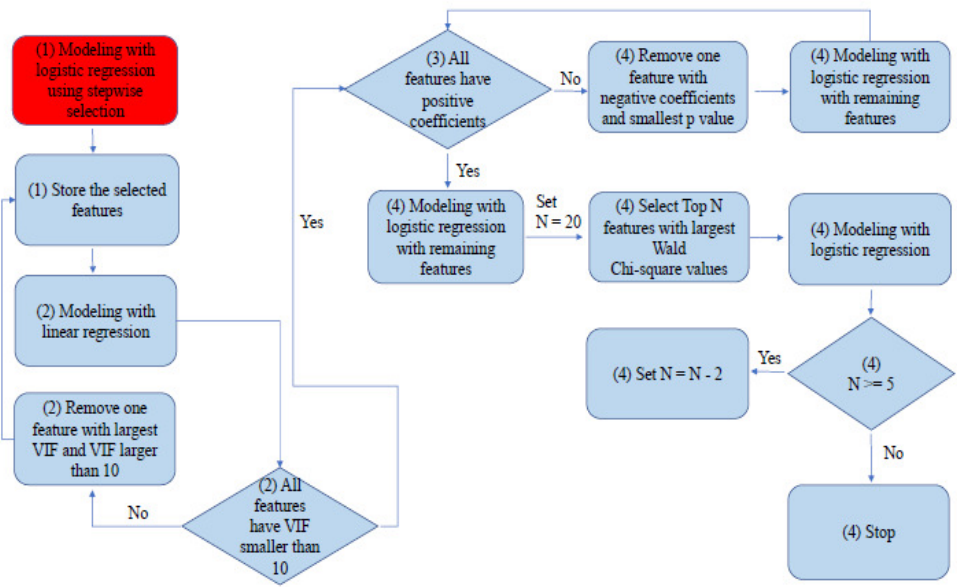

Figure 3. The block diagram of the second stage of the hybrid model.The labels (1), (2), (3), and (4) inside the diagram map to the steps in Section 3.4.

(4) Model optimization. This procedure is performed by watching the changing of model performance through gradually reducing the number of variables used in the model. Variables with the smallest absolute values of Wald Chi-square statistics (corresponding to the largest $p$ value) are first to be removed. In credit research area, the number of variables used for the model is preferred to be around 10. Therefore, in this paper, we first studied the model performance by using different number of variables. Then the model with relatively high ROC and $KS$ statistic on validation data while relative low number of variables would be recommended as the final model.

### 3.5. THE ONE-STAGE MODEL

To show the effectiveness of the hybrid model, or more specifically, the effectiveness of the newly created features by neural networks, we use the logistic regression without the neural network features as the baseline model. The logistic regression still follows the steps illustrated in Figure 3. We will call it the one-stage model in the remaining of this paper.

The difference between the one-stage model and the proposed two-stage hybrid model is that, the former uses the 178 features from Section 3.2 as predictors for model building while the latter uses the above 178 features plus the newly created features from Section 3.3. By comparing the performances of the two types of models, the effectiveness of the newly created features by using neural networks can be identified. Furthermore, the superiority of the proposed two-stage hybrid model over the one-stage model can also be demonstrated.

### 3.6. PERFORMANCE EVALUATION

In order to evaluate the performances of different models, model evaluation measures including the classification accuracy, Area Under the Curve ($AUC$), and $KS$ test were applied [28]. Denote True Positive (TP) as the customers with response that are correctly identified, False Positive (FP) as the customers without response that are identified as respondents, True Negative (TN) as the customers without response that are correctly identified, and False Negative (FN) as the customers with response that are identified as non-respondents. Then the classification accuracy could be defined in (6).

$$\text{Accuracy} = \frac{TP + TN}{TP + TN + FP + FN} \tag{6}$$

The second evaluation measure used in the paper is the $AUC$, where the curve is the receiver operating characteristic curve (ROC), which shows the interaction between the true positive rate (TPR, depicted in (7)) and the false positive rate (FPR, depicted in (8)) [29]. Greater $AUC$ denotes a better classification performance of the classifier.

$$\text{TPR} = \frac{TP}{TP + FN} \tag{7}$$

$$\text{FPR} = \frac{FP}{TN + FP} \tag{8}$$

The last evaluation measure applied is $KS$ test. The $KS$ statistic $D$ is defined in (9):

$$D = \max_s \left| F_n(s) - F_p(s) \right| \tag{9}$$

where $F_n(s)$ and $F_p(s)$ denotes the cumulative density function ($CDF$) of the classifier scores $= m(x)$ for negatives and positives, respectively. The purpose of $KS$ test is to use $D$ to test the null hypothesis that $CDF$ of negatives and positives are equivalent [30]. The value of $D$ indicates the furthest point on ROC curve from the diagonal (0, 0) to (1, 1) and larger value indicates better performance of the classifier[31].

# 4. EXPERIMENTAL RESULTS AND DISCUSSION BASED ON ATLANTICUS DATA

## 4.1. NEW FEATURES FROM NEURAL NETWORKS

We followed the block diagram in Figure1for new feature constructions by using neural network. As mentioned in Section3.3, in step D of Figure1,each of the 50 pairs of variables are used to build an individual neural network model. Each of these 50 neural networks is then used to obtain the predictions (denoted as$\hat{y}_0$, $\hat{y}_1$, ..., $\hat{y}_{49}$) of $RESP\_DV =$ 1 in step E. To demonstrate the construction of these predictions,$\hat{y}_1$, which is constructed based on variables *AMS3726* and *AMS3161*in this study, will be used as an example. The obtained neural network structure with weights and bias estimations is demonstrated in Figure 4. To further understand the relationships among *AMS3726*, *AMS3161*, and $\hat{y}_1$, the mathematical equation about how to calculate $\hat{y}_1$ based on *AMS3726* and *AMS3161* is shown inequation (10).

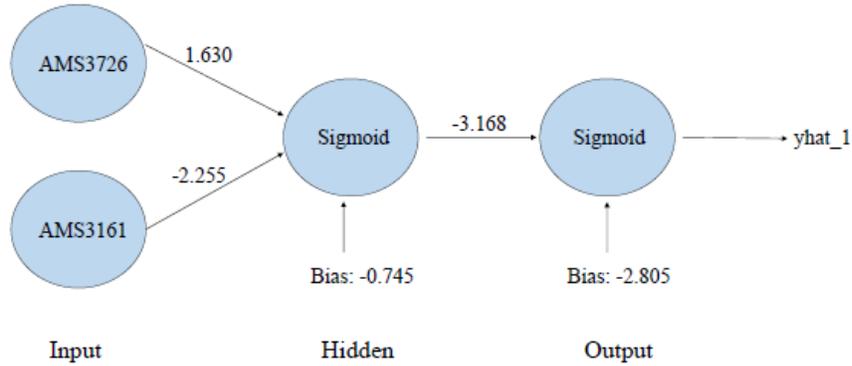

Figure 4. Illustration of the creation of $\hat{y}_1$

$$Z1 = -0.745 \ + \ 1.630 * \text{AMS3726} - 2.255 * \text{AMS3161}$$
$$A1 = sigmoid \ (Z1)$$
$$Z2 = -3.168 * A1 - 2.805 \qquad (10)$$
$$\hat{y}_1 = A2 = sigmoid \ (Z2)$$

As mentioned in Section 3.3, hierarchical variable clustering is performed on these 50 predictions (denoted as $\hat{y}_0$, $\hat{y}_1$, ..., $\hat{y}_{49}$) in step F of Figure 1 to get the final list of the new features. Figure 5 shows the result of hierarchical variable clustering analysis. With around 90%variations in the data are explained (thered vertical line in Figure 5), these 50 predictions form 22 clusters. Within each cluster, the variable with the lowest $1 - R^2 ratio$is then selected as the representative of the current cluster. As a result, 22 predictions (denoted as $\hat{y}_1$, $\hat{y}_3$,$\hat{y}_7$, $\hat{y}_9$,$\hat{y}_{12}$, $\hat{y}_{14}$,$\hat{y}_{15}$, $\hat{y}_{18}$,$\hat{y}_{19}$, $\hat{y}_{21}$,$\hat{y}_{25}$, $\hat{y}_{26}$,$\hat{y}_{27}$, $\hat{y}_{29}$,$\hat{y}_{30}$, $\hat{y}_{35}$,$\hat{y}_{37}$, $\hat{y}_{39}$,$\hat{y}_{43}$, $\hat{y}_{44}$, $\hat{y}_{46}$,$\hat{y}_{49}$)are selected as the representatives of the 50 predictions and are considered as the final newly constructed features.

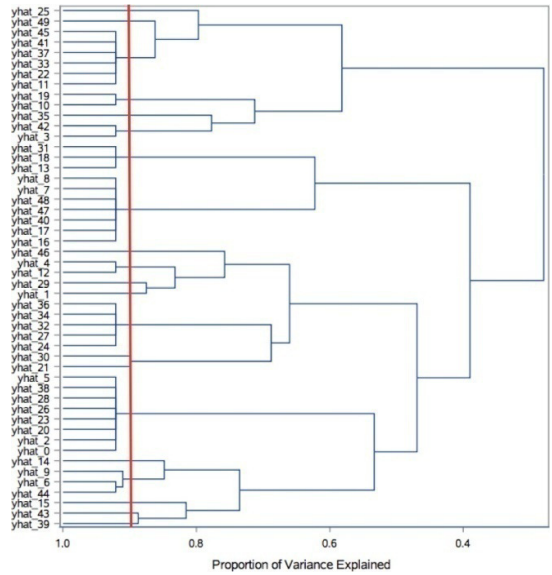

Figure 5. Hierarchical variable clustering to obtain newly created features

## 4.2. RESULTS OF THE TWO-STAGE HYBRID MODEL

The proposed two-stage hybrid model (logistic regression with newly created features by using neural network algorithm) was built by following the block diagram in Figure 3. Initially, 178 predictors plus the 22 newly created features from Section 4.1 (in total 200 features) were used as the input variables. Then, the full model (without feature selection) as well as eight other two-stage models with different number of features selected through the process in Figure 3 were built for bankcard response classifications. Table 1 shows the results of classification accuracy, $AUC$, and $KS$ in both training and validation sets based on the series of two-stage models. Again, all the variables used in Table 1 are already transformed to the WOE format.

With respect to Table 1, the full model always has the best performance with respect to classification accuracy, $AUC$, and $KS$ statistics due to making the best use of all the 200 features. As expected, the model performance with respect to classification accuracy, $AUC$, and $KS$ statistics show a non-increasing trend when the number of features decreases. Model 8 in Table 1(the model with six features selected) will be used as an illustrative example to demonstrate the modeling results of the two-stage model. Its coefficient estimations with corresponding $p$ values as well as the descriptions of the selected features are summarized in Table 2. It is observed that all the selected six features are highly significant in predicting the status of the customers. More over, they all have positive coefficient estimates in Table 2, which is consistent with the assumption that WOE-formed variables and thetarget variable have positive relationships.Our study also shows that the selected six features all have VIF values less than10(result not shown). Therefore, model 8 in Table 1 is considered as one of the optimal two-stage models. Its model function could be defined in equation (11), where $\hat{p}$ denotes the predicted probability of respondents (i.e., $RESP\_DV$ =1).

It is no table that, in (11), three newly created features ($\hat{y}_1$, $\hat{y}_{26}$, and $\hat{y}_{44}$) were selected as the significant features by model 8 in Table 3. This is strong evidence showing that the newly created features have significantly predictive power on the target variable. It is also reasonable to conclude that the new feature constructions by using neural networks in Section 3.3 is necessary.

Table 1. Performance of the two-stage model based on Atlanticus data. # of features denotes the number of features used in the model. Acc. denotes accuracy.

| Model Index | # of Features | *Acc.* on Train | *Acc.* on Valid | *AUC* on Train | *AUC* on Valid | *KS* on Train | *KS* on Valid |
|---|---|---|---|---|---|---|---|
| **Full Model** | 200 | 0.846 | 0.831 | 0.847 | 0.816 | 0.529 | 0.471 |
| **1** | 20 | 0.840 | 0.830 | 0.825 | 0.801 | 0.504 | 0.457 |
| **2** | 18 | 0.836 | 0.823 | 0.824 | 0.801 | 0.504 | 0.457 |
| **3** | 16 | 0.836 | 0.825 | 0.822 | 0.800 | 0.503 | 0.455 |
| **4** | 14 | 0.835 | 0.825 | 0.820 | 0.800 | 0.496 | 0.452 |
| **5** | 12 | 0.833 | 0.824 | 0.818 | 0.800 | 0.493 | 0.451 |
| **6** | 10 | 0.831 | 0.823 | 0.814 | 0.792 | 0.484 | 0.449 |
| **7** | 8 | 0.827 | 0.822 | 0.809 | 0.790 | 0.467 | 0.447 |
| **8** | 6 | 0.823 | 0.817 | 0.801 | 0.787 | 0.458 | 0.442 |

Table 2. Features selected by model 8 in Table 1.

| Feature Code | Estimate | *p Value* | Feature Label |
|---|---|---|---|
| Intercept | -9.244 | <0.001 | Model intercept |
| AMS3027 | 0.655 | <0.001 | Number of inquiries within 1 month |
| $\hat{y}_1$ | 3.374 | <0.001 | Newly created feature using AMS3726 and AMS3161 |
| AMS3124 | 0.556 | <0.001 | Age newest bankcard account |
| AMS3855 | 0.511 | <0.001 | Percent balance to high credit open department store accounts |
| $\hat{y}_{26}$ | 2.792 | <0.001 | Newly created feature using AMS3242 and AMS3193 |
| $\hat{y}_{44}$ | 4.061 | <0.001 | Newly created feature using AMS3828 and AMS3188 |

$$\log\left(\frac{\hat{p}}{1-\hat{p}}\right) = -9.244 + 0.655 * AMS3027 + 3.374 * \hat{y}_1$$
$$+0.556 * AMS3124 + 0.511 * AMS3855$$
$$+2.792 * \hat{y}_{26} + 4.061 * \hat{y}_{44}$$

(11)

Table 3. Performance of the one-stage modelbased on Atlanticus data. # of features denotes the number of features used in the model. Acc. denotes accuracy.

| Model Index | # of Features | *Acc.* on Train | *Acc.* on Valid | *AUC* on Train | *AUC* on Valid | *KS* on Train | *KS* on Valid |
|---|---|---|---|---|---|---|---|
| **Full Model** | 178 | 0.841 | 0.827 | 0.845 | 0.802 | 0.499 | 0.441 |
| **1** | 20 | 0.834 | 0.825 | 0.825 | 0.792 | 0.499 | 0.438 |
| **2** | 18 | 0.834 | 0.823 | 0.824 | 0.792 | 0.493 | 0.439 |
| **3** | 16 | 0.833 | 0.823 | 0.821 | 0.790 | 0.490 | 0.431 |
| **4** | 14 | 0.831 | 0.822 | 0.819 | 0.787 | 0.486 | 0.426 |
| **5** | 12 | 0.830 | 0.824 | 0.817 | 0.785 | 0.479 | 0.421 |
| **6** | 10 | 0.825 | 0.824 | 0.804 | 0.775 | 0.474 | 0.415 |
| **7** | 8 | 0.821 | 0.823 | 0.800 | 0.768 | 0.458 | 0.415 |
| **8** | 6 | 0.815 | 0.815 | 0.777 | 0.756 | 0.417 | 0.395 |

It is worth to mention that, model 8 in Table 1 is not the only satisfying two-stage models based on the dataset used in this study. According to the modeling regulation and criterion in the financial institutions, the number of features used in the final model should not be too large (usually around 10). Therefore, models 5, 6, 7, and 8 in Table 1 are all candidate two-stage models for bankcard response classification purpose. It is because the performances of the above four models do not have too much decrease when compared with the full model while they are using much fewer features. However, because of using different datasets in the bankcard response tasks, it is risky to make general conclusions on the optimal models. Future researchers can refer to the workflow shown in this paper as a guide for making decisions on final optimal models.

### 4.3. RESULTS OF THE ONE-STAGE MODEL

As discussed in Section 3.5, the one-stage model ignores the first stage in the hybrid model but still follows the block diagram in Figure 3. The 178 predictors following the data pre-processing were used as the input variables into the one-stage model. As a result, the full model (without feature selection) as well as eight other one-stage models with different number of features were built for bankcard response classifications. The results for classification accuracy, $AUC$, and $KS$ statistics in both training and validation sets are demonstrated in Table 3.

Again, the full model always has the best performance with respect to classification accuracy, $AUC$, and $KS$ statistics due to making the best use of the information provided by all the 178 variables. We still see that the model performance with respect to classification accuracy, $AUC$, and $KS$ statistics show a non-increasing trend when the number of features decreases. Similar with Table 2, we summarize the results of the one-stage model 8 in Table 4 by giving the $\beta$ estimates, the corresponding $p$ value, and the variable labels. Once more, all the selected six features are highly significant in predicting the status of the customers with positive coefficient estimates. The estimated equation is given in (12).

Table 4. Features selected by model 8 in Table 3.

| Feature Code | Estimate | p Value | Feature Label |
|---|---|---|---|
| Intercept | -13.613 | <0.001 | Model intercept |
| AMS3027 | 0.761 | <0.001 | Number of inquiries within 1 months |
| AMS3726 | 0.811 | <0.001 | Number open bankcard accounts with update within 3 months |
| AMS3215 | 1.413 | <0.001 | Number accounts with past due amount > 0 |
| AMS3855 | 0.669 | <0.001 | Percent balance to high credit open department store accounts |
| AMS3828 | 0.942 | <0.001 | Percent revolving accounts to accounts |
| AMS3124 | 0.474 | <0.001 | Age newest bankcard account |

$$\log\left(\frac{\hat{p}}{1-\hat{p}}\right) = -13.613 + 0.761 * AMS3027 + 0.811 * AMS3726$$
$$+1.413 * AMS3215 + 0.669 * AMS3855$$
$$+0.942 * AMS3828 + 0.474 * AMS3124$$

(12)

Similar to the results from the two-stage models, there is no standard answer for the best one-stage model based on different datasets and different modeling tasks. But the work flow provided in this study could be used as a reference for future researchers in dealing with bank card response problems.

### 4.4. MODEL COMPARISON

By comparing results summarized in Tables 1 and 3, it can be concluded that, the proposed two-stage hybrid model in general has a better performance than the one-stage model in terms of the classification accuracy, $AUC$, and $KS$ statistics when the same number of features are selected. Since $KS$ statistics measures the degree of separation between the positive and negative distributions in the dataset, it is weighted more than classification accuracy and $AUC$ in the bank card response classification in this study. From the results inTables1and3, we can conclude that the two-stage hybrid model has much better performance interms of $KS$ statistics on validation sets incomparison with that of one-stage model. For example, the two-stage model 8 from Table1selects six features and can achieve the $KS$ statistics valued 0.442 on the validation set. This is about 12%increasecomparedtothevalue0.395basedontheone-stagemodel8from Table 3 with the same number of features selected.

Some researchers may argue that, as the result shown in Table 2, three of the six selected features in the two-stage model 8 from Table 1are the newly created features based on other independent variables, meaning that there are actually nine features used bythismodel.Tomakethecomparisonfair,wehavefurtherfittheone-stage model with nine features selected. As a result, it can achieve the $KS$ statistics valued 0.415 on the validation set. Thus, we are confident enough to conclude that the proposed two-stage model has better differentiable capability between positive and negatives in terms of $KS$ statistics compared to one-stage model when the same number of features are used.

Another view of Table 1 shows that, even though the two-stage model 8 uses only six features (or nine features as mentioned above), the obtained $KS$ statistics valued 0.442 on validation set is still higher than that from the one-stage full model valued 0.441 from Table 3. Since the $KS$ statistics on validation set in Table 1 shows a decreasing trend with the decreasing number of features used, it is reasonable to say that the two-stage model 8 in Table 1 has a better performance than all the one-stage models in Table 3. The newly created features by using neural network algorithms in the first stage of the hybrid model are shown to be a good support for identifying complex relationships among variables. Consequently, we can conclude that the proposed two-stage hybrid model outperforms the commonly utilized one-stage model and hence provides efficient alternatives in conducting bankcard response tasks.

## 5. FURTHER MODEL EVALUATION BASED ON PUBLIC HMEQ DATA

To further confirm the consistency, stability and reliability of the proposed two-stage hybrid model, the public dataset HMEQ [10] (available in the SAMPSIO library of SAS and also at http://www.creditriskanalytics.net/)is used. The HMEQ dataset describes whether the applicant has defaulted on the home equity line of credit. It contains records from 5,960 applicants, and 12 features that are related with the clients' credit information. The target variable $BAD$ indicates whether an applicant defaulted on his/her loans and the default rate in the dataset is 80.05%.To use this dataset in this study, the categorical values of the features have been transformed to numerical values.

For the HMEQ dataset, the methods for data pre-processing, neural networks for new feature construction, one-stage and two-stage modeling, as well as the performance evaluation are all the same as those used for the Atlantic us data. After data pre-processing, the training set has 3,577 records while the validation set has 2,383 records with 11 independent variables remain. These 11 variables form 55 possible pairs by using the $n$-$choose$-$k$ combination described in (3). Thus, 55 different logistic regressions with 1-way interactions were built in step A described in Figure 1, which took about 20 minutes in SAS on the same computer as that for the Atlanticus data. Then in step C, the value of $N$ was set to 6 for HMEQ data via trying different values ranging from 5 to 50. Six different neural networks were built in step D in Python, which took less than 2 seconds. Finally, in steps E and F, 4 newly created features were identified as additional predictors.

Tables 5 and 7 show the results fromthe two-stage and the one-stage model based on the HMEQ data, respectively. It is observed that when using the same number of features, the two-stage model has a better performance than the one-stage model with respect to classification accuracy, $AUC$ and $KS$ statistics. This result is consistent with that based on theAtlanticus data. Tables 6 and 8 show the $\beta$ estimates, the corresponding $p$ value, and the variable labels for the 4th two-stage modeland 3rd one-stage model based on HMEQ data, respectively. It is notable that in the 4th two-stage model, one newly created feature $\hat{y}_0$was selected as the significant feature. This further confirms the necessity of the new feature construction stage in the proposed hybrid model. Last but not the least, model 4 of the two-stage model (with 5 features, or, 6 features if researchers argue that $\hat{y}_0$ was created based on 2 original features) even has better performance than model 2 of the one-stage model (with 7 features). This makes us more confident about the better performance of the two-stage model compared with the one-stage model.

Table 5. Performance of the two-stage model based on HMEQ data. # of features denotes the number of features used in the model. Acc. denotes accuracy.

| Model Index | # of Features | *Acc.* on Train | *Acc.* on Valid | *AUC* on Train | *AUC* on Valid | *KS* on Train | *KS* on Valid |
|---|---|---|---|---|---|---|---|
| **Full Model** | 15 | 0.840 | 0.843 | 0.815 | 0.800 | 0.483 | 0.468 |
| **1** | 11 | 0.840 | 0.843 | 0.815 | 0.798 | 0.473 | 0.459 |
| **2** | 9 | 0.838 | 0.838 | 0.791 | 0.780 | 0.446 | 0.431 |
| **3** | 7 | 0.836 | 0.831 | 0.790 | 0.781 | 0.436 | 0.429 |
| **4** | 5 | 0.835 | 0.830 | 0.787 | 0.775 | 0.430 | 0.413 |

Table 6. Features selected by model 4 in Table 5.

| Feature Code | Estimate | *p Value* | Feature Label |
|---|---|---|---|
| Intercept | -5.682 | <0.001 | Model intercept |
| $\hat{y}_0$ | 4.545 | <0.001 | Newly created feature using LOAN and MORTDUE |
| DELINQ | 0.622 | <0.001 | Number of delinquent credit lines |
| DEBTINC | 0.068 | <0.001 | Debt-to-income ratio |
| JOBLEVLE | 0.145 | <0.001 | Newly created variable to indicate occupational categories |
| NINQ | 0.162 | <0.001 | Number of recent credit inquiries |

Table 7. Performance of the one-stage model based on HMEQ data. # of features denotes the number of features used in the model. Acc. denotes accuracy.

| Model Index | # of Features | *Acc.* on Train | *Acc.* on Valid | *AUC* on Train | *AUC* on Valid | *KS* on Train | *KS* on Valid |
|---|---|---|---|---|---|---|---|
| **Full Model** | 11 | 0.838 | 0.836 | 0.799 | 0.792 | 0.448 | 0.443 |
| **1** | 9 | 0.833 | 0.835 | 0.777 | 0.782 | 0.434 | 0.428 |
| **2** | 7 | 0.830 | 0.831 | 0.766 | 0.769 | 0.400 | 0.409 |
| **3** | 5 | 0.831 | 0.830 | 0.757 | 0.755 | 0.386 | 0.404 |

Table 8. Features selected by model 3 in Table 7.

| Feature Code | Estimate | p Value | Feature Label |
|---|---|---|---|
| Intercept | -4.695 | <0.001 | Model intercept |
| DEROG | 0.623 | <0.001 | Number of major derogatory reports |
| DELINQ | 0.650 | <0.001 | Number of delinquent credit lines |
| NINQ | 0.157 | <0.001 | Number of recent credit inquiries |
| DEBTINC | 0.063 | <0.001 | Debt-to-income ratio |
| JOBLEVEL | 0.134 | <0.001 | Newly created variable to indicate occupational categories |

# 6. CONCLUSIONS AND AREAS OF FUTURE RESEARCH

In the financial domain, more and more companies are seeking better strategies for decision making through the help of bankcard response models. Hence the bankcard response models have drawn serious attention during the past decade. Logistic regression and LDA are the most commonly utilized statistical techniques in the credit research domain. However, these techniques only focus on exploring linear relationship among variables and sometimes produce poor bankcard response capabilities. In this situation, the neural network, which could handle the nonlinear relationship among the variables, represents a powerful and attractive choice in dealing with bankcard response problems due to its outstanding classification capability. However, in the meanwhile, neural network is being criticized for its long training process, limited ability to magnitude the variable importance, complex topological structure, as well as no well-established criteria for the interpretations of the coefficients. Furthermore, due to the regulations and policies in financial institutions, logistic regression is widely acceptable while neural networks have very limited applicability as classification or prediction tools.

In this paper, we focus on making full use of the advantages of the neural network while avoid its disadvantages. The purpose is to propose a two-stage hybrid approach by using neural network as a feature construction tool(instead of a classification or prediction tool) to improve the performance of bankcard response model. The rationale underlying the analyses is firstly using the neural networks to create new features. Since neural networks could identify the underlying online are relationship between variables, the newly created features are supposed to contribute to the success of the subsequent model building tasks. Then in the second stage, the newly created features are added as additional input variables in logistic regression.

To demonstrate the effectiveness of the proposed two-stage hybrid bankcard response model, its performance is compared with that from the one-stage model (without using the neural networks to create new features) after applied to the Atlantic us data using holdout cross validation approach. The results demonstrate that by identifying new features, the hybrid two-stage modelling general out performs the one-stage model interms of classification accuracy, *AUC* and *KS* statistics. By checking the two-stage model with six features selected, it is found that three of these six features are the new features created by the neural network algorithm. This could further confirm the effectiveness of the feature construction step in the two-stage model. Finally, the public HMEQ data was used to further evaluate the reliability of the proposed model. As the result shows, the same conclusions can be made based on the HMEQ data, which could further confirm the consistency and stability of the proposed two-stage method.

Compared to the previous studies summarized in Section 2, the two-stage hybrid model proposed in this paper have many advantages:

(1) Different from many previous studies that use the neural network as a classification or prediction tool, we use it as a feature construction tool. The newly created features could denote the non linear relationships among variables. In the meanwhile, the neural network structure used in the proposed model is very simple. This can overcome the short comings of the neural network in terms of its complex topology and limited interpretability.

(2) When using neural network as the feature construction tool in this study, only the subset of the dataset is used. This could reduce the training time in comparison with building neural networks on the entire data set, thus overcome the short comings of the neural network in terms of its long processing time when the dataset is relatively large.

(3) Due to the regulation or policy restrictions in the financial institutions, logistic regression is the only acceptable tool for classifications or predictions in many cases. The two-stage model in this study demonstrates the capability of neural networks in creating new while important features and hence can improve the performance of logistic regression. Therefore, the framework proposed in this paper provide efficient alternatives for future researchers in conducting bankcard responseproblems.

To improve the accuracy of bankcard response model, many researchers have tried to explore the important variables in their modeling procedure by using the feature selection algorithms. Therefore, in future studies, the effectiveness of the proposed two-stage model can be compared with modeling based on some feature selection algorithms, such as simulated annealing [32], F-score LDA [33], and particle swarm optimization [34]. Moreover, except neural network algorithms, it is possible to use other classification techniques (including discriminant analysis, bagging and boosting algorithms, decision tree, and support vector machine) as feature construction tools. As another recommendation, the proposed model in this paper can be used on other data sets to evaluate its generalizability.

## ACKNOWLEDGEMENTS

The authors would like to thank Atlantics Services Corporation (located at Atlanta, GA, USA) for providing the credit customer response data set.

## AUTHORS

**Yan Wang** is a Ph.D. candidate in Analytics and Data Science at Kennesaw State University. Her research interest contains algorithms and applications of data mining and machine learning techniques in financial areas. She has been a summer Data Scientist intern at Ernst & Young and focuses on the fraud detections using machine learning techniques. Her current research is about exploring new algorithms/models that integrates new machine learning tools into traditional statistical methods, which aims at helping financial institutions make better strategies. Yan received her M.S. in Statistics from University of Georgia.

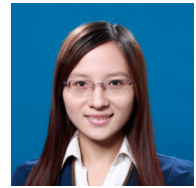

**Dr.Xuelei Sherry Ni** is currently a Professor of Statistics and Interim Chair of Department of Statistics and Analytical Sciences at Kennesaw State University, where she has been teaching since 2006. She served as the program director for the Master of Science in Applied Statistics program from 2014 to 2018, when she focused on providing students an applied leaning experience using real-world problems. Her articles have appeared in the Annals of Statistics, the Journal of Statistical Planning and Inference and StatisticaSinica, among others. She is the also the author of several book chapters on modeling and forecasting. Dr.Ni received her M.S. and Ph.D. in Applied Statistics from Georgia Institute of Technology.

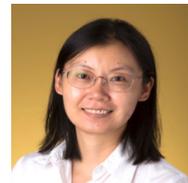